%% file: main.tex
\documentclass[10pt,twocolumn,letterpaper]{article}

\usepackage{cvpr}              %
\usepackage[accsupp]{axessibility}  %

\usepackage{graphicx}
\usepackage{amsmath}
\usepackage{amssymb}
\usepackage{mathabx}
\usepackage{booktabs}
\usepackage{algorithm}
\usepackage{algpseudocode}

\usepackage{xpatch}
\makeatletter
\xpatchcmd{\algorithmic}
  {\ALG@tlm\z@}{\leftmargin\z@\ALG@tlm\z@}
  {}{}
\makeatother

\usepackage[pagebackref,breaklinks,colorlinks]{hyperref}

\usepackage[capitalize]{cleveref}
\crefname{section}{Sec.}{Secs.}
\Crefname{section}{Section}{Sections}
\Crefname{table}{Table}{Tables}
\crefname{table}{Table}{Tables}

\usepackage{multirow}
\usepackage[table,xcdraw]{xcolor}

\definecolor{osa}{HTML}{E0E0E0} %
\definecolor{pspm}{HTML}{C0C0C0} %

\usepackage{enumitem}
\newcommand{\squishlist}{
   \begin{list}{$\bullet$}
    { \setlength{\itemsep}{0pt}      \setlength{\parsep}{3pt}
      \setlength{\topsep}{3pt}       \setlength{\partopsep}{0pt}
      \setlength{\leftmargin}{1.0em} \setlength{\labelwidth}{1em}
      \setlength{\labelsep}{0.5em} } }
      
\newcommand{\squishend}{
    \end{list}  }

\newlength\replength
\newcommand\repfrac{.33}

\newcommand\rulewidth{.6pt}
\newcommand\tdashfill[1][\repfrac]{\cleaders\hbox to \replength{%
  \smash{\rule[\arraystretch\ht\strutbox]{\repfrac\replength}{\rulewidth}}}\hfill}

\newcommand\tdotfill[1][\repfrac]{\cleaders\hbox to \replength{%
  \smash{\raisebox{\arraystretch\dimexpr\ht\strutbox-.1ex\relax}{.}}}\hfill}

\usepackage{nicefrac}
\usepackage{amsthm}

\theoremstyle{remark}

\theoremstyle{definition}

\usepackage[margin=0pt,position=bottom,font=footnotesize,labelfont=bf]{caption}

\newcommand{\PP}[1]{\paragraph*{\bf #1}}

\def\cvprPaperID{11148} %
\def\confName{CVPR}
\def\confYear{2022}

\title{Zero-Query Transfer Attacks on Context-Aware Object Detectors}
\author{Zikui Cai\textsuperscript{\rm 1}, Shantanu Rane\textsuperscript{\rm 2}, Alejandro E. Brito\textsuperscript{\rm 2},  Chengyu Song\textsuperscript{\rm 1}, \\ Srikanth V. Krishnamurthy\textsuperscript{\rm 1}, Amit K. Roy-Chowdhury\textsuperscript{\rm 1}, M. Salman Asif\textsuperscript{\rm 1 \thanks{Corresponding author}}\\
\textsuperscript{\rm 1}University of California, Riverside, USA~~~~
\textsuperscript{\rm 2}Palo Alto Research Center, Palo Alto, USA\\
{\tt\small \{zcai032\}@ucr.edu, \{srane, abrito\}@parc.com} \\
{\tt \small  \{csong, krish\}@cs.ucr.edu, \{amitrc, sasif\}@ece.ucr.edu}
}

\begin{document}

\maketitle

\begin{abstract}
Adversarial attacks perturb images such that a deep neural network produces incorrect classification results. A promising approach to defend against adversarial attacks on 
natural multi-object scenes is to impose a context-consistency check, wherein, if the detected objects are not consistent with an appropriately defined context, then an attack is suspected. Stronger attacks are needed to fool such context-aware detectors. We present the first approach for generating context-consistent adversarial attacks that can evade the context-consistency check of
black-box object detectors operating on complex, natural scenes. Unlike many black-box attacks that perform repeated attempts and open themselves to detection, we assume a ``zero-query'' setting, where the attacker has no knowledge of the classification decisions of the victim system.
First, we derive multiple attack plans that assign incorrect labels to victim objects in a context-consistent manner. Then we design and use a novel data structure that we call the perturbation success probability matrix, which enables us to filter the attack plans and choose the one most likely to succeed. This final attack plan is  implemented using a perturbation-bounded adversarial attack algorithm. We compare our zero-query attack against a few-query scheme that repeatedly checks if the victim system is fooled. We also compare against state-of-the-art context-agnostic attacks. Against a context-aware defense, the fooling rate of our zero-query approach is significantly higher than context-agnostic approaches and higher than that achievable with up to three rounds of the few-query scheme.  
\end{abstract}

\input{introduction}
\input{backgrounds}
\input{method}

\input{experiment}

\input{discussion}

\section{Conclusions}
In this paper, we craft a novel zero-query attack by exploiting ``a context graph'' that captures co-occurrence relations of objects in a natural image. Against context-aware detectors, the fooling rate is significantly higher than that achieved by a context-agnostic attack. Unlike prior query-based attacks, our attack is extremely hard to detect since it hinges on using a single attempt. It achieves fairly good fooling rates by choosing an attack plan (i.e., perturbing multiple objects simultaneously to ensure context consistency) which is likely to succeed. 
The key innovation is a PSPM that provides this information offline. We observe that the use of PSPM not only boosts fooling rates in white-box settings, but also carries over to the black-box setting (i.e., when the detector  model is different from that of the attacker) consistently for different attacker-defender pairs.

\noindent\textbf{Acknowledgments.} 
This material is based upon work supported by the Defense Advanced Research Projects Agency (DARPA) under agreement number HR00112090096.
Approved for public release; distribution is unlimited.

{\small
\bibliographystyle{ieee_fullname}
\bibliography{refs}
}

\input{supp}

\end{document}

%% file: introduction.tex
\section{Introduction}

Despite achieving significant performance gains on a variety of vision and language tasks, deep neural networks (DNNs) are vulnerable to adversarial attacks \cite{szegedy2013intriguing}. One of the most popular adversarial approaches is the class of perturbation-bounded evasion attacks \cite{goodfellow2014explaining,papernot2017practical,carlini2017towards,madry2017towards}. Here, an attacker can make a model yield arbitrarily wrong classification results by adding imperceptible perturbations to the input image. These attacks are quite practical and can be performed at test time without needing access to the training data.
The vast majority of work in this area has focused on attacking classifiers trained on datasets like ImageNet, MNIST, CIFAR-10, and CIFAR-100, where the classifier attempts to recognize one dominant object in a given image. 
\begin{figure}[t]
    \centering
    \includegraphics[width=\linewidth]{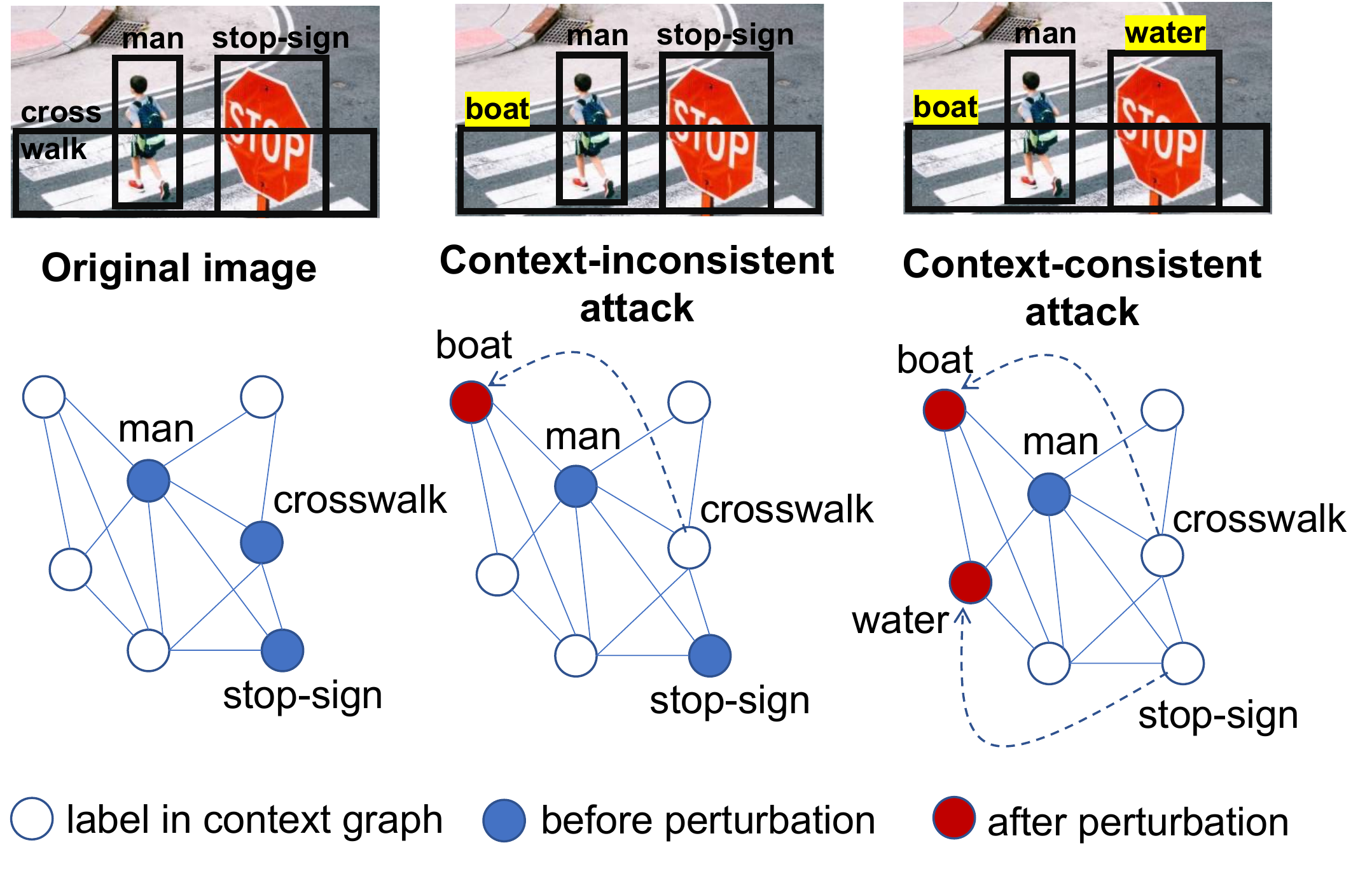}
    \caption{ For natural scenes containing multiple objects, applying an evasion attack on an individual object (e.g., crosswalk $\rightarrow$ boat) violates the context: A boat and a stop-sign rarely occur together. A context-aware detector can detect this attack. In this work, we perturb multiple objects in a context-consistent way (e.g., crosswalk $\rightarrow$ boat, stop-sign $\rightarrow$ water) in a single attempt. The combination (person, boat, water) does not violate context and thus fools even a context-aware detector.}
    \label{fig:consistency-check}
\end{figure}
In contrast, we are primarily concerned with object detectors \cite{ren2015faster,Redmon2016YOLO,lin2017focal,kong2020foveabox,pang2019libra}
that localize and recognize multiple objects in an image, which is the case in most natural images.
Such detectors often take a holistic view of an image, rather than considering it as a collection of arbitrary objects~\cite{xie2017adversarial,wu2020making}.
The objects in natural scene images form a context that can help identify the scene or given the scene we are likely to find the objects that conform to the scene context. For instance, a boat is unlikely to co-occur with a stop sign, and much more likely to co-occur with water. Leveraging this observation, some recent attack \cite{cai2022context} and defense \cite{li2020connecting,yin2021exploiting}  mechanisms have been proposed that take image context into account.  
Context-aware defense methods in \cite{li2020connecting,yin2021exploiting} can detect attacks that are inconsistent with the scene context, as shown in \Cref{fig:consistency-check}. To evade these defenses, changing one object in the image is insufficient. 
The context-aware attack method in \cite{cai2022context} uses the knowledge of co-occurrence between different object classes in complex images to generate a sequence of transferable attacks for black-box object detectors; however, this method needs a few queries to test which attack plan is successful. 
\emph{We present, for the first time, a zero-query attack algorithm that changes multiple objects simultaneously in a context-consistent manner thereby creating a holistic adversarial scene that can overcome context-aware defenses.}

In this work, we consider ``zero-query'' attacks (ZQA) that refer to a setting in which the attacker has no feedback channel to access the classification decisions of the victim system. This setting is extremely useful in practice because in many applications the victim system is inaccessible to the attacker; even if the victim system is accessible, the attacker's communications can be monitored, and thus draw suspicion. ZQA, on the other hand, is a truly stealthy attack.
The attacker can only implement an attack plan once by perturbing multiple objects in a given scene and submitting the perturbed image to the victim system. Furthermore, we assume that the victim system is explicitly context-aware; that is, it will examine the list of detected objects and determine whether that list is ``context-consistent'' or not. If yes, then the detector will not suspect an attack. If not, it will suspect that the image has been perturbed by an attacker. Our ZQA approach is able to subvert more sophisticated multi-label object detectors that either implicitly or explicitly take context relationships across objects into account while performing their inference. In fact, accounting for context is what makes it possible to achieve high success rates in a single attempt. 

 Several approaches exist for scene context modeling~\cite{hollingworth1998does,oliva2003top,torralba2003contextual,dvornik2018modeling,barnea2019exploring}. In this paper, we restrict our attention to object co-occurrence, which is the most fundamental approach to modeling semantic context. The context model is represented by a co-occurrence graph (or equivalently the co-occurrence matrix) that is computed for a given set of images. We consider a list of objects as context-consistent only if the corresponding labels form a fully connected sub-graph within the co-occurrence graph.

The main contributions of this paper are as follows.
\squishlist

\item We develop an architecture for designing attacks on multi-object scenes that fool context-aware object detectors. Our detectors explicitly use object co-occurrence to model scene context.

\item We propose an approach for zero-query context-aware attacks that generate adversarial scenes to fool a context-aware detector in a single step. 

\item %
We introduce the concept of a perturbation success probability matrix (PSPM) that models the probability of successfully perturbing a given target object to a given victim object in the white box setting. We use the PSPM to refine our attack plans, essentially choosing the one which is most likely to succeed. We show that the PSPM-guided attacks improve the fooling rate even in a black-box setting. %

\item %
We show experimentally that the fooling rate of ZQA is significantly higher than that achieved by a context-agnostic black-box attack.  
Furthermore, we compare our results against a possible ``few-query'' strategy~\cite{cai2022context} that repeatedly enhances the attack plan, while observing the detector output, until the detector is fooled. For the Pascal VOC dataset~\cite{everingham2010pascal}, the ZQA attacks provide fooling rates comparable to 5-query and 3-query attacks in the white-box and black-box settings, respectively. 
\squishend

%% file: backgrounds.tex
\section{Background and Preliminaries}

\subsection{Context in Object Detection}
The role of context in improving visual recognition tasks has been well studied~\cite{mottaghi2014role,Salakhutdinov2011multiclass,Yao2012WholeScene}. The state-of-the-art object detectors locate and detect several objects in the scene based on holistic information in the image~\cite{ren2015faster,Redmon2016YOLO}. Many of these methods explicitly utilize  context information to improve the performance of object detectors~\cite{Bell2016ION,chen2018context,liu18cvpr,Barnea2019ConBound,Beery2020CRCNN,Wang2020Occlu}. Co-occurrence-based contextual information derived from image pixels~\cite{li2020connecting}, object labels \cite{cai2022context}, and language models~\cite{yin2021exploiting} have also been used to build a context-aware object detector. Different from the above contributions, to the best of our knowledge, this is the first paper that proposes a context-aware attack to fool a context-aware detector with zero queries.

\subsection{Black-Box Attacks}
In a black-box attack, the attacker has no access to the internal parameters of the model; thus, instead of generating the perturbed image by backpropagation, the attacker can only test a perturbed image on the victim model. In some cases, the attacker can observe the output but in many cases even that is not possible \cite{li2020practical}. This renders query-based attacks inapplicable, which usually take an overwhelmingly large number (often hundreds or thousands) of  queries~\cite{brendel2017decision,chen2017zoo,guo2019simple,chen2020hopskipjumpattack,li2020qeba,Wang2020blackobject}. 
In this paper, we explore the most stringent case where no model queries are allowed. Such attacks will be extremely hard to detect, free from suspicion of repeated queries, and thus will be a more viable option for subversion. Several papers \cite{papernot2017practical,Liu2017Delving,dong2018boosting,li2020towards} have examined the phenomenon of transfer attacks where the adversarial examples generated using a surrogate network can fool a black-box victim network. A large body of work exists on designing (perturbation-bounded) evasion attacks for images containing one predominant object and evaluating how well they transfer in a black-box setting \cite{goodfellow2014explaining,carlini2017adversarial,papernot2016transferability,chakraborty2018adversarial,silva2020opportunities}. In this work, our goal is to fool object detectors for general scenes. This is considered a harder problem because of the need to perturb multiple objects~\cite{xie2017adversarial,wu2020making}. This difficulty is exacerbated by the need to preserve contextual consistency during the attack~\cite{li2020connecting}. A ``few-query'' strategy for context-aware attacks was proposed in \cite{cai2022context} that repeatedly enhances the attack plan, while observing the detector output, until the detector is fooled. Our goal in this paper is to develop context-aware attacks with zero queries.

\subsection{Attacks against Object Detectors} 
Attacking object detectors is harder than attacking classifiers, since the attack must obfuscate the category as well as the location of one or more objects~\cite{xie2017adversarial,wu2020making}. Object detectors~\cite{ren2015faster,Redmon2016YOLO} implicitly use contextual information -- for instance, relationships between object pixels and background pixels -- to increase the speed of inference. Researchers have exploited this fact by developing various kinds of adversarial patches~\cite{Liu2019Dpatch,Saha2020SPatch,Hu2020CCA} which do not overlap with the victim objects, but can still fool the detectors. Some other attacks \cite{xie2017adversarial,Zhang2020CAA,Wei2019Transferable,chow2020adversarial}, perturbing the image globally, are also successful in fooling object detectors in a white-box setting. A recently proposed method has demonstrated the ability to transfer attack black-box object detectors by exploiting context information, but the victim models do not explicitly check for context-consistency and also the attack needs multiple queries \cite{cai2022context}.

\subsection{Defense methods}
Some representative defense mechanisms \cite{ren2020adversarial} for mitigating adversarial attacks include enhancing adversarial robustness of the model intrinsically through adversarial training \cite{madry2017towards,tramer2017ensemble,bai2021recent} or strengthening model architectures \cite{liu2018towards,xie2019feature}; and destroy adversary externally through input transformation \cite{guo2017countering,xie2017mitigating} or denoising \cite{xu2017feature,samangouei2018defense,meng2017magnet}. These defenses are context-agnostic. Some recent papers consider context-aware object detectors operating on natural multi-object scenes \cite{li2020connecting,yin2021exploiting}. Although these works use different notions of context from our work, they confirm that attacking a context-aware detector is more difficult. 

%% file: method.tex
\section{Context-Aware Zero-Query Attacks}

\begin{figure*}
    \centering
    \includegraphics[width=0.85\linewidth]{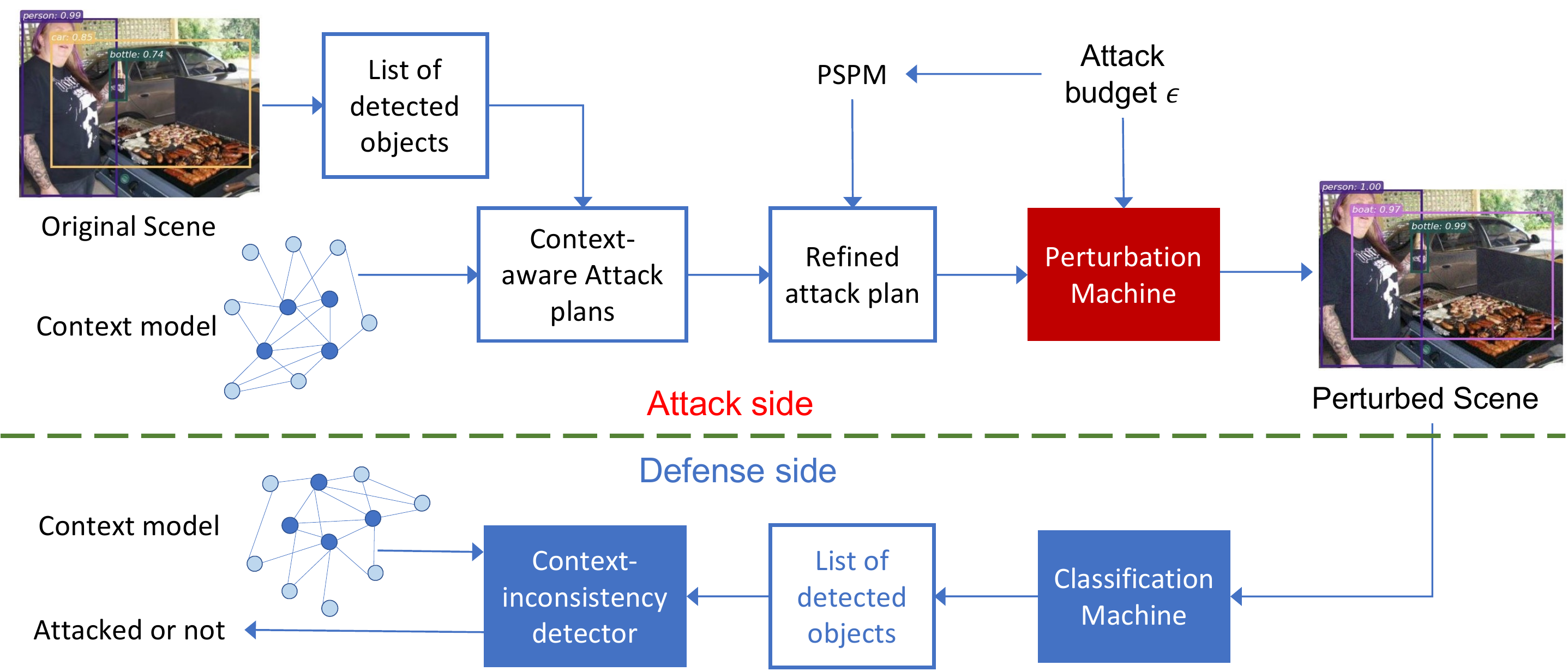}
    \caption{High-level diagram of zero-query context-aware black-box attacks. Given a victim image to be attacked, the attacker first finds a list of detected objects in the image and then consults the semantic context associated with the detector to design a context-aware attack plan that perturbs a  victim object to a target label. To improve attack success rate, the attacker checks the PSPM corresponding to the perturbation machine with a certain perturbation budget $\epsilon$ and refines the original attack plan. With the refined attack plan, the attacker perturbs the image within bound $\epsilon$, using the perturbation machine. The attacker's action is now complete. The perturbed image is then sent to a black-box classification / detection machine equipped with an explicit context-consistency detection mechanism. The attack is considered successful only if the victim object is successfully perturbed to the target object and the context-inconsistency detector does not find any inconsistency in the list of all detected objects.}
    \label{fig:overview}
\end{figure*}

We describe ZQA for natural scenes.
A high-level diagram of zero-query context-aware black-box attacks is shown in~\Cref{fig:overview}. We first present a high-level description and later elaborate on the building blocks. The attacker determines a list of objects detected in the scene and uses the co-occurrence-based context model to derive several context-aware attack plans that perturb one or more target objects to their respective victim labels. Then, given the perturbation budget, the attacker refines the list using a pre-computed PSPM, which we will discuss in detail later. The result is an attack plan that consists of a list of (victim label, target label) pairs that are most likely to succeed in fooling the victim object detector. The attacker then uses an evasion attack algorithm to generate the perturbed scene according to the refined attack plan. 
The perturbed image is sent to a black-box classification / detection machine equipped with an explicit context-consistency detection mechanism. The attack is considered successful only if the victim object is successfully perturbed to the target label  \emph{and} the victim system's object detector does not find any context inconsistency in its list of detected objects. We now describe the building blocks of the attack in detail.

\subsection{Context Model}
\label{subsec:anomaly-detector}

The context model is used by the attacker and the victim system's object detector to determine whether a given list of objects is context-consistent or not. Let us first define what is considered as context-consistent and what is context-inconsistent. Intuitively, combinations of objects detected in natural images from the training data should be considered as context-consistent, since %
these objects appear together in such scenes. On the other hand, combinations of objects should be considered as context-inconsistent if some of the objects have never appeared together in the training data. Thus, object co-occurrence is a fundamental cue in determining context-consistency. 

\begin{figure}
    \centering
    \includegraphics[width=2.75in]{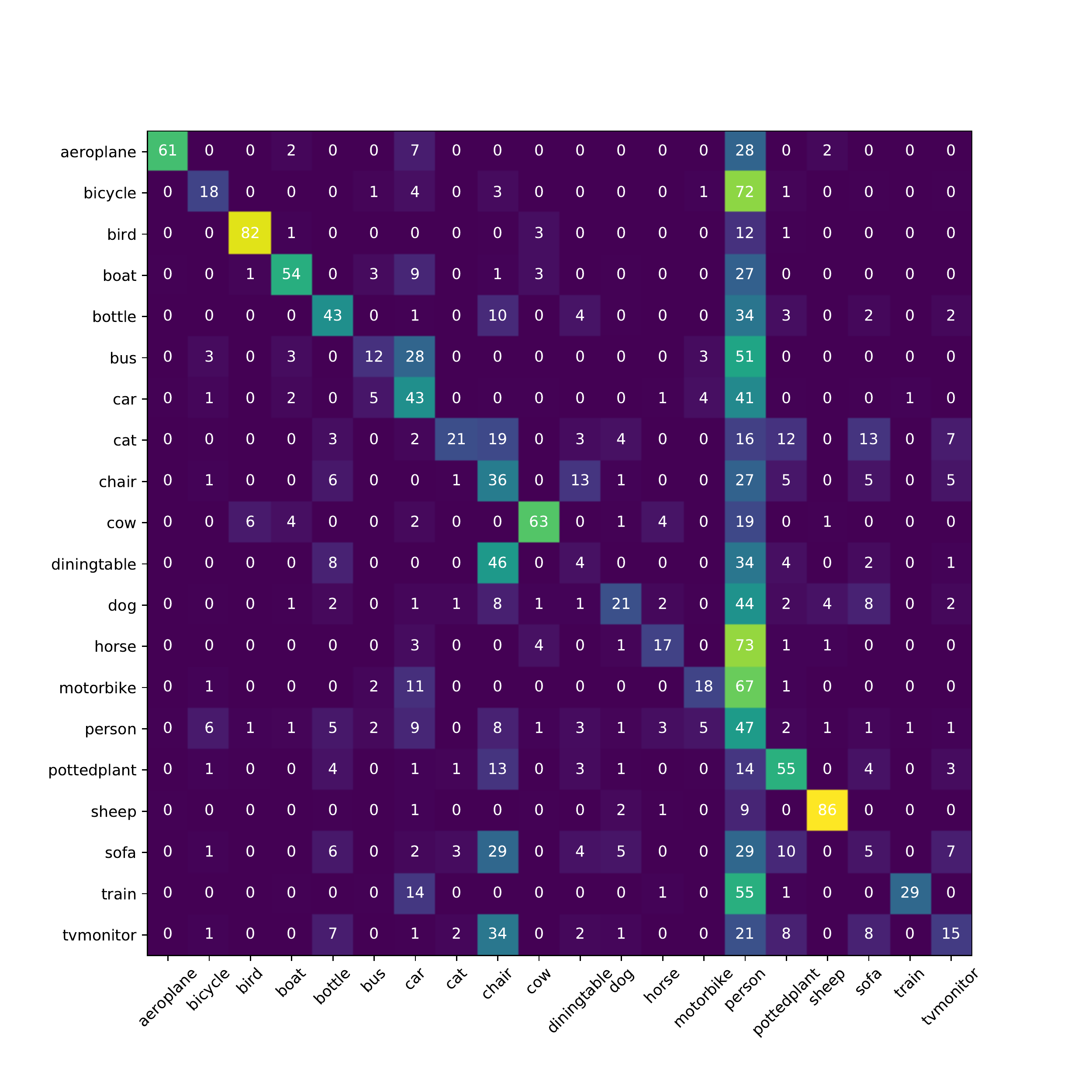}
    \caption{Co-occurrence matrix (conditional probability form) for the Pascal VOC07 training data set. Each cell indicates the probability as a quantized integer percentage (\%).}
    \label{fig:co-occurrence-matrix}
\end{figure}

\bigskip 
\noindent\textbf{Co-occurrence matrix/graph.} We build a matrix, called the co-occurrence matrix $\mathbf{G}$, to model the co-occurrence relationships between objects as follows. 
Given a training data set with $N$ labels and a label set $\mathcal{N} = \{ \ell_1, \ldots, \ell_N \}$, a co-occurrence matrix $\mathbf{G}$ is an $N \times N$ matrix whose entries $\mathbf{G}(i,j)$ represent the number of unique pairs of objects with labels $\ell_i, \ell_j \in \mathcal{N}$ appearing together in the images. This matrix is symmetric before normalization. After normalizing each entry in $\mathbf{G}$ with the sum of elements in its row, we obtain $\widebar{\mathbf{G}}$, where $\widebar{\mathbf{G}}(i,j)$  indicates the conditional probability $p_{j|i}$ which is the probability of observing label $\ell_j$ if label $\ell_i$ is observed. In general, $\widebar{\mathbf{G}}$ is not symmetric. An example of $\widebar{\mathbf{G}}$ for the Pascal VOC Dataset is illustrated in \Cref{fig:co-occurrence-matrix}. 
The co-occurrence matrix can also be interpreted as a context graph (which we will also denote by $\mathbf{G}$) with $N$ nodes, where the weight of the edge between nodes $i$ and $j$ represents the number of times that those two labels appear together in the training data. 

\bigskip 
\noindent\textbf{Context consistency.}
If two nodes in a context graph do not have an edge connecting them, then these two labels never appear together in an image. Using this notion of co-occurrence-based context, we can define context-consistency and context-inconsistency as follows. 
A list of objects is considered as \textit{context-consistent} if all the nodes representing the labels of those objects form a fully connected subgraph of the context graph ${\mathbf{G}}$. All the natural images in the data set satisfy this condition.
A list of objects is considered as \textit{context-inconsistent} if there are at least two nodes, representing the labels of two objects, that do not have an edge between them in the context graph.

\bigskip 
\noindent \textbf{Generalized context-consistency.}
Suppose we threshold the entries of ${\mathbf{G}}$ to obtain a matrix  ${\mathbf{H}}_{\eta}$, where ${\mathbf{H}}_{\eta}(i,j) = {\mathbf{G}}(i,j)$ if ${\mathbf{G}}(i,j) > \eta$, for some threshold $\eta$. Otherwise ${\mathbf{H}}_{\eta}(i,j) = 0$. Build a context graph using ${\mathbf{H}}_{\eta}$, which we will also denote by the same symbol ${\mathbf{H}}_{\eta}$. Then the co-occurrence based notion of context consistency can be readily generalized as follows. 
A list of objects is considered as context-consistent up to a threshold $\eta$ if all the nodes representing the labels of those objects form a fully connected subgraph of the context graph ${\mathbf{H}}_{\eta}$. Generalized context-inconsistency is defined similarly.

We describe the notion of context consistency for a list of object labels in \Cref{fig:consistency-check}. Let us take a clean image containing three objects as an example. Suppose we have a person, a crosswalk and a stop-sign in the image as indicated by the dark blue nodes. These three nodes form a fully connected subgraph indicating that these three objects are context consistent. Suppose we want to perturb the crosswalk in the image to a boat. An example of a context-inconsistent attack involves just perturbing the victim object (crosswalk) to the target label (boat denoted as a red node). The perturbed object list (person, boat, stop-sign) is no longer context-consistent because these three nodes of boat, person and stop-sign do not form a fully connected graph, indicating that the combination never appears in the training data.  Intuitively, it would be unlikely to see a boat with a stop-sign in the natural images. An example of a context-consistent attack is to perturb the  crosswalk to a boat but also perturb the stop-sign to water. That combination (person, water, boat) does appear in the training data, as seen from the fact that the label nodes form a fully connected sub-graph.

\subsection{Perturbation Success Probability Matrix}
\label{subsec:pspm}
Perturbation Success Probability Matrix (PSPM) is an $N \times N$ matrix denoted as $\mathbf{M}_{\mathcal{C},\epsilon, \alpha}$ that is defined for an ensemble of classification models $\mathcal{C}$,  perturbation budget $\epsilon$,  and an object perturbation algorithm $\alpha$. PSPM is defined for a specific training data set with labels $\mathcal{N} = \{\ell_1, \ldots, \ell_N \}$.
$\mathbf{M}_{\mathcal{C},\epsilon, \alpha}(i,j)$ encodes the probability that an object with label $\ell_i$ can be successfully perturbed to label $\ell_j$ by the perturbation algorithm $\alpha$ using the perturbation budget $\epsilon$.

\begin{figure}
    \centering
    \includegraphics[width=2.75in]{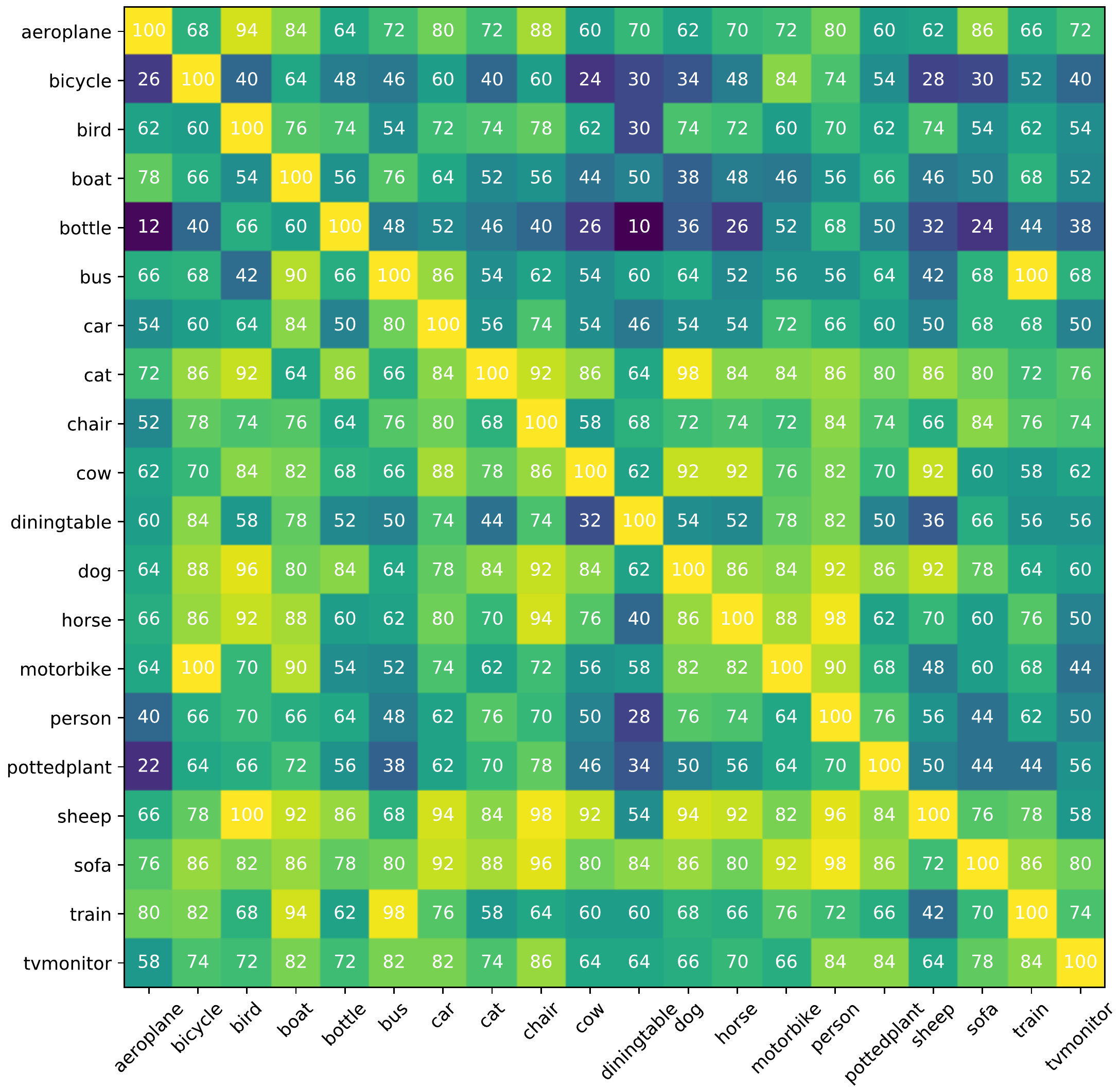}
    \caption{PSPM of Pascal VOC07 data set for $\mathcal{C} = \{\text{``Faster R-CNN''}\}$, $\epsilon = 10$, $\alpha = \{``\text{PGD}'' \}$.  Each cell indicates the probability as a quantized integer percentage (\%).}
    \label{fig:pspm}
\end{figure}

The PSPM expresses an attacker's ability to perturb  \emph{individual} objects in a scene. 
The utility of the PSPM can be explained as follows. Suppose the list of objects in the given scene is $\mathcal{A} = \{A_1, \ldots, A_S\}$. Suppose the list $\mathcal{A}$ is perturbed to a list $\mathcal{B}$ using an evasion attack algorithm $\alpha$ with perturbation budget $\epsilon$. A context-agnostic attack would choose the labels in $\mathcal{B}$ at random from the label set $\mathcal{N}$. A context-aware attack would make sure that the labels in $\mathcal{B}$ are context-consistent as determined by the co-occurrence matrix $\widebar{\mathbf{G}}$. Then, there are in general one or more possible perturbation assignments $\mathcal{A} \rightarrow \mathcal{B}$ that are context-consistent. Depending upon the training set (e.g., the presence of objects in different poses, sizes, illuminations) some object perturbations, $A_i \rightarrow B_j$, are likely to be more successful than others, even in a white-box setting. Thus, each perturbation assignment $\mathcal{A} \rightarrow \mathcal{B}$ suggested by an examination of the co-occurrence matrix has a different likelihood of success. The PSPM enables us to select the assignment that is most likely to succeed. An example of a PSPM for the Pascal VOC07 dataset is shown in \Cref{fig:pspm}.

Some ways to choose an assignment amongst many assignments permitted by the co-occurrence matrix include
\begin{enumerate}
    \item Choose the assignment $\mathcal{A} \rightarrow \mathcal{B}$ that maximizes each  $\mathbf{M}_{\mathcal{C},\epsilon,\alpha}(i,j)$ with $i \in \mathcal{A}, j \in \mathcal{B}$.
    \item Choose the assignment $\mathcal{A} \rightarrow \mathcal{B}$ that maximizes the average of all $\mathbf{M}_{\mathcal{C},\epsilon,\alpha}(i,j)$ with $i \in \mathcal{A}, j \in \mathcal{B}$.
    \item Choose the assignment $\mathcal{A} \rightarrow \mathcal{B}$ that maximizes the minimum of all $\mathbf{M}_{\mathcal{C},\epsilon,\alpha}(i,j)$ with $i \in \mathcal{A}, j \in \mathcal{B}$.
\end{enumerate}

For simplicity, we use the first approach in the ensuing development. Admittedly, the PSPM considers the success of perturbing some $A_i \in \mathcal{A}$ to $B_j \in \mathcal{B}$ in a white-box setting; that is, for one or more classification models specified in the given ensemble $\mathcal{C}$. In a black-box setting, the attacker does not know which model is being used at the victim side. Thus, attack plans based on the PSPM are, at best, approximations. We hypothesize that these approximations are still better than choosing, at random, one of many possible assignments $\mathcal{A} \rightarrow \mathcal{B}$ suggested by the co-occurrence matrix. Our experiments, described in Section~\ref{sec:expts}, indicate that refining the attack plan using the PSPM indeed improves the fooling rate compared to choosing an attack plan at random from the context-consistent candidate attacks suggested by the co-occurrence matrix.

\subsection{Context-Consistent Attack Plan Generation}

We now describe the zero-query method for deriving context-consistent attack plans using the co-occurrence matrix $\widebar{\mathbf{G}}$ and the PSPM matrix $\mathbf{M}_{\mathcal{C},\epsilon,\alpha}$. The overall procedure is described in \cref{alg:oneshot}. We assume that there is a desired target assignment for one object in the scene (e.g., change one of the horses to a bicycle). As discussed, just perturbing one object can result in a context-inconsistent list of objects in the perturbed scene. Thus, other objects in the scene may need to be perturbed so that the resulting list is context-consistent. The procedure described below ensures not only the attack plan is context-consistent, but it is also the plan that is most likely to succeed in a white-box setting. As we mentioned earlier, for a black-box setting this plan may not always be the most likely to succeed. Experimentally, we found that the attack plan guided by PSPM also increases the fooling rate for black-box models.

\begin{algorithm}
\caption{Zero-Query Context-Consistent Attack Plan}\label{alg:oneshot}
\begin{algorithmic}
\\
\noindent 
\textbf{Inputs}: 
A list of objects in the scene $\mathcal{A}= \{A_1=\ell_v, A_2, \ldots, A_S \}$ where $S$ denotes the total number of objects in the scene,  the desired target assignment consisting of (victim label $\ell_v \in \mathcal{A}$, target label $\ell_t$) pair for one object, 
co-occurrence matrix $\mathbf{G}$ computed over the training data, co-occurrence threshold $\eta$, perturbation budget $\epsilon$, perturbation success probability matrix $\mathbf{M}_{\mathcal{C},\epsilon,\alpha}$ generated in advance from the training data.
\vspace{0.1in}
\\
\noindent
\textbf{Output}: Attack plan $\mathcal{B}$ that is context-consistent with $\mathbf{G}$ and most likely to succeed in a white-box setting.
\vspace{0.1in}
\\
\noindent
\textbf{Initialization}: $\mathcal{B} = \{ \ell_t \}$
\vspace{0.1in}
\\
\noindent
\textbf{Procedure}: \\
\begin{enumerate}
    \item Obtain a set of labels that co-occur with $\ell_t$, denoted as $\mathcal{T} \subset \mathcal{N}$ such that ${\mathbf{G}}(j,t) > \eta$ for all $\ell_j \in \mathcal{T}$.
    \item Set running counter $k = 2$.
    \item While $k \leq S$, compute
\begin{align}
\ell^* ~ &= ~ \underset{\ell_b \in \mathcal{T}}{\operatorname{argmax}}~\mathbf{M}_{\mathcal{C},\epsilon,\alpha}(a,b),  \label{eq:osa-pspm} \\ 
&\text{where $a$ is the index of label}~A_k \notag \\
\mathcal{B} &\gets \mathcal{B} \bigcup \ell^* \notag \\
k &\gets k+1 \notag
\end{align}
\end{enumerate}
\end{algorithmic}
\end{algorithm}

\subsection{Implementation of Attack Plan}\label{sec:generate_attacks}
To generate the adversarial scene, evasion attacks can be implemented using a single or multiple surrogate model(s). Our attack generation method with a single surrogate detector is based on projected gradient descent (PGD) \cite{madry2017towards} within a $\ell_\infty$ ball, which can be considered as a powerful multi-step variant of FGSM \cite{goodfellow2014explaining}. We initialize a zero perturbation $\delta^0 = 0$, and update it in each iteration as
\begin{equation}\label{eq:PGD}
    \delta^{t+1} = \Pi_{\mathcal{S}} \left( \delta^{t} - \lambda~ \text{sgn}(\nabla_{\delta}\mathcal{L}(x+\delta^{t}, y)) \right),
\end{equation}
where $\mathcal{L}$ is the loss function, $x \in \mathbb{R}^d$ is the input image and $y$ is the target label. We generate the desired output $y$ based on our attack plan, which includes object categories, locations, and confidence scores. We take a step size $\lambda$ at each iteration $t$, and project (clip) $\Pi_{\mathcal{S}}$ the perturbation $\delta^{t}$ to the feasible set $\mathcal{S}$ which satisfies the following two criteria
\begin{equation}
    \begin{cases}
      \hspace{0.5cm} \| \delta^{t} \|_\infty \leq \epsilon, \\
      \hspace{0.5cm} x + \delta^{t} \in [0,255]^d.
    \end{cases}       
\end{equation}

We use PGD for its simplicity in our experiments, but we can easily modify our method to use other (more advanced) perturbation methods such as MIM~\cite{dong2018boosting} and DIM~\cite{xie2019improving} etc. without losing generalizability.

%% file: experiment.tex
\section{Experiments}
\label{sec:expts}

We performed extensive experiments on two large-scale object detection datasets to evaluate the proposed context-aware zero-query attack strategy. We construct a query-based baseline scheme
and evaluate the fooling rate of our proposed zero-query approach against it.

\PP{Datasets:}
We conduct our experiments using images from both PASCAL VOC~\cite{everingham2010pascal} and MS COCO~\cite{lin2014microsoft} datasets. VOC contains 20 common classes of objects, and COCO contains 80 classes which is a super-set of the categories in VOC. We randomly selected 500 images from \texttt{voc2007test} and \texttt{coco2017val} respectively, which contain multiple $(2-6)$ objects. This manuscript contains results for various models on the \texttt{voc2007test}. The results for the \texttt{coco2017val} are in the supplementary.

\PP{Attack models:}
To mimic a realistic black-box setting, we pick a variety of object detectors, including two-stage detectors: Faster RCNN \cite{ren2015faster} and Libra R-CNN \cite{pang2019libra}; one-stage detector: RetinaNet~\cite{lin2017focal}; and anchor-free detector: FoveaBox~\cite{kong2020foveabox}.
We use implementations of the aforementioned models from the \texttt{MMDetection} \cite{mmdetection} code repository. The models in \texttt{MMDetection} are trained on \texttt{coco2017train}; therefore, while testing the detectors on VOC images, we only return the object labels  available in VOC. The models under such adaptation still get good detection performance, as shown in \Cref{tab:map}.

\begin{table}[h]
\centering
\caption{Mean average precision (mAP) at IOU (intersection over union) threshold 0.5 of different detectors used in our experiments. Models are evaluated on VOC07 test set. \textbf{Legend:} Faster R-CNN (FRCNN), RetinaNet (Retina), Libra R-CNN (Libra), FoveaBox (Fovea).}
\label{tab:map}
\begin{tabular}{ccccc}
\hline
Model   & FRCNN & Retina & Libra & Fovea \\
mAP@.50 & 78.30\%      & 78.51\%   & 79.01\%     & 77.68\%  \\ \hline
\end{tabular}
\end{table}

\PP{Zero-Query attacks (ZQA and ZQA-PSPM):}
We evaluate two variants of our zero-query attack. The first variant (ZQA) ensures that the attack plan is chosen at random from the available set of context-consistent attacks and is determined using \Cref{eq:osa}. The second variant (ZQA-PSPM) generates a context-consistent attack plan based on the PSPM matrix that was pre-computed for the given classification model, perturbation budget and evasion attack algorithm. See \Cref{eq:osa-pspm}. ZQA-PSPM is the key contribution of this paper, and our results demonstrate that that ZQA-PSPM provides better fooling success rate compared to ZQA and baselines.

\PP{Baselines and comparisons:}
We compare the ZQA and ZQA-PSPM schemes against two relevant baselines. The first is the context-agnostic zero-query attack, which we call ``Context-Agnostic''. In this attack, the attack plan that drives the scene perturbation is chosen randomly (i.e., without explicitly enforcing co-occurrence-based context). This means that some attack plans in the Context-Agnostic scheme may be context-consistent by accident, while others are context-inconsistent. Comparison against Context-Agnostic is performed with the aim of investigating the benefits of exploiting context while designing the attack plan.

We also compare with a second, more powerful baseline, which we refer to as the ``Few-Query'' approach \cite{cai2022context}. In this scheme, the attacker is equipped with the co-occurrence matrix ${\mathbf{G}}$ but doesn't have the PSPM matrix. More importantly, the few-query attacker can query the victim system to find out whether the attack succeeded. Because of this, they don't need to perturb all the objects in the scene in one step. The few-query attacker proceeds as described in Algorithm~\ref{alg:fewshot}. The few-query attack is denoted as ``Few-Query $q$" in \cref{tab:compare-frcnn} and \cref{tab:compare-libra} where $q$ is the number of previous queries that the current attack is built on, $q \in \{0,1,2,3,4,5\}$, thus ``Few-Query $0$" is identical to ZQA in terms of queries.

\begin{algorithm}
\caption{Few-Query Context-Consistent Attack Plans}\label{alg:fewshot}
\begin{algorithmic}
\\
\noindent 
\textbf{Inputs}: Same as the inputs to Algorithm~\ref{alg:oneshot} but with no PSPM matrix, number of victim system queries allowed $q \le S$ where $S$ is the number of objects in the scene.
\vspace{0.1in}
\\
\noindent
\textbf{Output}: A sequence of attack plans $\mathcal{D}_k$ consistent with ${\mathbf{G}}$, where $k = 1, 2, ..., q$.
\vspace{0.1in}
\\
\noindent
\textbf{Initialization}: $\mathcal{D}_1 = \{ \ell_t \}$
\vspace{0.1in}
\\
\noindent
\textbf{Procedure}: 
\begin{enumerate}
    \item Obtain the label set $\mathcal{T} \subset \mathcal{N}$ such that ${\mathbf{G}}(j,t) > \eta$ for all $\ell_j \in \mathcal{T}$ .
    \item Set running counter $k = 2$. 
    \item While $k \leq q$, compute
\begin{align}
&\widehat{\ell} ~ \overset{iid}{\sim} ~ U(\mathcal{T}) \quad \rhd \text{uniformly sample from}~ \mathcal{T} \label{eq:osa}\\
&\mathcal{D}_k \gets \mathcal{D}_{k-1} \bigcup \widehat{\ell} \notag \\
&k \gets k+1 \notag
\end{align}
\end{enumerate}
The attacker implements $\mathcal{D}_1$ and queries the victim system. If the attack succeeded, they stop; else,  they implement $\mathcal{D}_2$ and query the victim system, and so on, until the attack succeeds. 
\end{algorithmic}
\end{algorithm}

\PP{Attack generation:}
We use the PGD-based method to generate a perturbation on the whole image (as discussed in \Cref{eq:PGD}). We experiment with $L_{\infty}$ perturbation budget $\epsilon \in \{10,20,30,40,50\}$. The step size $\lambda = 2$, and maximum number of iterations is $50$. We observe that when an object is very close to or overlaps with the victim object, perturbing that object to any label different from the victim's target label reduces the success rate; thus, we map all objects whose regions have $\text{IOU} \ge 0.3$ with the victim object to the victim's target label.

\input{tables/compare_frcnn}
\input{tables/compare_libra}

\PP{Evaluation metrics:}
We use the metric  ``context-consistent attack success rate'' (or fooling rate) to evaluate the  attack performance on a victim object detector. For an attack to be regarded as a successful context-consistent attack, it must (1) successfully perturb the victim object to the target label, and (2) pass the context-consistency check described in~\Cref{fig:consistency-check}. We define the fooling rate as the percentage of the number of test cases for which the above two conditions are satisfied.

\PP{Comparing zero-query attacks with few-query attacks:}
The fooling rates of the few-query attacks with different numbers of queries (Few-Query 0 to Few-Query 5) and the zero-query attack under white-box and black-box settings at different perturbation budgets are shown in \cref{tab:compare-frcnn} and \cref{tab:compare-libra}. The settings for both tables are detailed in the captions. The
fooling rates for the few-query attacks are cumulative; that is, the values reported for few-query-$k$ accounts for successful attacks with $0, 1,\ldots, k$ queries. 

We observe that ZQA can achieve higher fooling rates than the few-query attack for up to 4 queries in the white-box setting and up to 2 queries in the black-box setting. When PSPM is used to refine the zero-query attack plan (ZQA-PSPM), the fooling rate increases, outperforming up to 5 queries of the few-query attack in the white-box setting and up to 3 queries in the black-box setting. These results are consistent across several detector models tested. As $\epsilon$ reduces, the perturbation is not always enough to carry out the evasion attacks, and thus the fooling rates fall from $\epsilon = 50$ to $\epsilon=10$.

The results clearly demonstrate the advantage of simultaneous context-aware perturbation of all objects in the scene. In many cases, using the PSPM to refine the context-aware attack further improves the fooling rate. While the few-query approach eventually outperforms the ZQA attack, recall that the former requires the attacker to communicate with the detector, which is either not always possible, or might expose the attacker.

%% file: tables/compare_frcnn.tex
\begin{table*}[ht]
\centering
\caption{
Fooling rates (\%) of different attack strategies under different $L_{\infty}$ perturbation $\le \epsilon \in \{50,40,30,20,10\}$. We compare ZQA and ZQA-PSPM with Context-Agnostic ZQA, and Few-Query attacks where feedback from blackbox (BB) models is allowed.  
The white-box (WB) is \textbf{Faster R-CNN} and three black-box models (BB1, BB2, BB3) are \textbf{RetinaNet, Libra R-CNN and FoveaBox} respectively. Fooling rate is counted as the percentage of attacks where victim is perturbed to a target label and all detected labels satisfy context consistency. Tested on 500 images from VOC 2007 test set which contain multiple (2-6) objects. Shaded cell indicates up to which few-query step, ZQA or ZQA-PSPM
has better performance than few-query attack. \colorbox{osa}{Lighter shades} are for ZQA, and \colorbox{pspm}{darker shades} are for ZQA-PSPM.}
\label{tab:compare-frcnn}
\scalebox{0.75}{
\begin{tabular}{c|cccccccccccccccccccc}
\toprule
\rowcolor[HTML]{FFFFFF} 
\cellcolor[HTML]{FFFFFF} & \multicolumn{4}{c}{\cellcolor[HTML]{FFFFFF}$\epsilon = 50$} & \multicolumn{4}{c}{\cellcolor[HTML]{FFFFFF}$\epsilon = 40$} & \multicolumn{4}{c}{\cellcolor[HTML]{FFFFFF}$\epsilon = 30$} & \multicolumn{4}{c}{\cellcolor[HTML]{FFFFFF}$\epsilon = 20$} & \multicolumn{4}{c}{\cellcolor[HTML]{FFFFFF}$\epsilon = 10$} \\ \cline{2-21} 
\rowcolor[HTML]{FFFFFF} 
\multirow{-2}{*}{\cellcolor[HTML]{FFFFFF}Method} & WB & BB1 & BB2 & BB3 & WB & BB1 & BB2 & BB3 & WB & BB1 & BB2 & BB3 & WB & BB1 & BB2 & BB3 & WB & BB1 & BB2 & BB3 \\ 
\midrule
\rowcolor[HTML]{FFFFFF} 
Context-Agnostic & 34.0 & 29.0 & 30.0 & 25.4 & 36.8 & 26.2 & 30.0 & 29.6 & 35.4 & 27.4 & 31.2 & 27.8 & 35.2 & 24.4 & 30.8 & 27.6 & 30.4 & 13.8 & 15.6 & 17.8 \\
\rowcolor[HTML]{FFFFFF} 
\cellcolor[HTML]{E0E0E0}ZQA & 90.0 & 46.6 & 52.2 & 54.0 & 92.0 & 48.0 & 55.0 & 51.8 & 91.6 & 46.0 & 57.0 & 52.2 & 87.4 & 39.6 & 50.4 & 51.0 & 65.2 & 21.0 & 23.8 & 24.2 \\
\rowcolor[HTML]{FFFFFF} 
\cellcolor[HTML]{C0C0C0}ZQA-PSPM & \textbf{92.6} & \textbf{51.2} & \textbf{61.6} & \textbf{56.8} & \textbf{92.0} & \textbf{51.8} & \textbf{55.4} & \textbf{54.4} & \textbf{93.0} & \textbf{49.2} & \textbf{57.2} & \textbf{54.0} & \textbf{88.2} & \textbf{44.0} & \textbf{51.4} & \textbf{53.4} & \textbf{70.6} & \textbf{23.2} & \textbf{27.4} & \textbf{28.2} \\ 
\hline \hline
\rowcolor[HTML]{E0E0E0} 
\cellcolor[HTML]{FFFFFF}Few-Query 0 & 60.0 & 29.8 & 29.8 & 34.8 & 64.2 & 34.6 & 34.2 & 39.8 & 66.2 & 34.2 & 35.0 & 37.8 & 61.2 & 29.2 & 30.8 & 35.8 & 48.2 & 14.8 & 14.0 & 20.2 \\
\rowcolor[HTML]{E0E0E0} 
\cellcolor[HTML]{FFFFFF}Few-Query 1 & 64.4 & 35.8 & 40.4 & 43.0 & 68.0 & 41.0 & 44.2 & 49.4 & 69.6 & 41.8 & 45.2 & 47.6 & 68.2 & 36.0 & 40.2 & 43.4 & 58.4 & \cellcolor[HTML]{C0C0C0}21.6 & 23.8 & \cellcolor[HTML]{C0C0C0}27.8 \\
\cellcolor[HTML]{FFFFFF}Few-Query 2 & \cellcolor[HTML]{E0E0E0}77.6 & \cellcolor[HTML]{C0C0C0}48.0 & \cellcolor[HTML]{C0C0C0}56.2 & \cellcolor[HTML]{FFFFFF}59.0 & \cellcolor[HTML]{E0E0E0}80.0 & \cellcolor[HTML]{C0C0C0}50.0 & \cellcolor[HTML]{E0E0E0}52.6 & \cellcolor[HTML]{FFFFFF}57.2 & \cellcolor[HTML]{E0E0E0}78.4 & \cellcolor[HTML]{C0C0C0}47.0 & \cellcolor[HTML]{E0E0E0}54.2 & \cellcolor[HTML]{FFFFFF}54.6 & \cellcolor[HTML]{E0E0E0}77.6 & \cellcolor[HTML]{C0C0C0}43.8 & \cellcolor[HTML]{E0E0E0}50.0 & \cellcolor[HTML]{E0E0E0}49.8 & \cellcolor[HTML]{C0C0C0}70.4 & \cellcolor[HTML]{FFFFFF}27.0 & \cellcolor[HTML]{FFFFFF}29.6 & \cellcolor[HTML]{FFFFFF}34.2 \\
\rowcolor[HTML]{FFFFFF} 
Few-Query 3 & \cellcolor[HTML]{E0E0E0}86.8 & 55.4 & 65.0 & 65.8 & \cellcolor[HTML]{E0E0E0}89.6 & 55.4 & 58.6 & 62.0 & \cellcolor[HTML]{E0E0E0}86.2 & 54.8 & 62.0 & 61.2 & \cellcolor[HTML]{E0E0E0}86.4 & 49.6 & 57.2 & 55.4 & 77.0 & 31.4 & 34.0 & 39.8 \\
\rowcolor[HTML]{FFFFFF} 
Few-Query 4 & \cellcolor[HTML]{C0C0C0}91.6 & 60.0 & 71.8 & 69.4 & 95.2 & 61.4 & 63.8 & 66.0 & \cellcolor[HTML]{E0E0E0}91.2 & 58.0 & 68.2 & 67.2 & 89.6 & 53.2 & 60.6 & 59.8 & 81.4 & 34.2 & 37.8 & 43.2 \\
\rowcolor[HTML]{FFFFFF} 
Few-Query 5 & 95.0 & 61.8 & 75.0 & 73.4 & 97.2 & 62.8 & 68.0 & 69.4 & 96.2 & 61.2 & 71.0 & 70.6 & 93.0 & 56.6 & 65.2 & 63.4 & 85.2 & 35.8 & 40.2 & 46.2 \\
\bottomrule
\end{tabular}}
\end{table*}

%% file: tables/compare_libra.tex
\begin{table*}[ht]
\centering
\caption{
Follow the setting in \Cref{tab:compare-frcnn} but use \textbf{Libra R-CNN} as WB and use \textbf{Faster R-CNN, RetinaNet and FoveaBox} as BB1, BB2, BB3 respectively.
Fooling rates (\%) of different attack strategies under different $L_{\infty}$ perturbation $\le \epsilon \in \{50,40,30,20,10\}$ are as follows.
}
\label{tab:compare-libra}
\scalebox{0.75}{
\begin{tabular}{c|cccccccccccccccccccc}
\toprule
\rowcolor[HTML]{FFFFFF} 
\cellcolor[HTML]{FFFFFF} & \multicolumn{4}{c}{\cellcolor[HTML]{FFFFFF}$\epsilon = 50$} & \multicolumn{4}{c}{\cellcolor[HTML]{FFFFFF}$\epsilon = 40$} & \multicolumn{4}{c}{\cellcolor[HTML]{FFFFFF}$\epsilon = 30$} & \multicolumn{4}{c}{\cellcolor[HTML]{FFFFFF}$\epsilon = 20$} & \multicolumn{4}{c}{\cellcolor[HTML]{FFFFFF}$\epsilon = 10$} \\ \cline{2-21} 
\rowcolor[HTML]{FFFFFF} 
\multirow{-2}{*}{\cellcolor[HTML]{FFFFFF}Method} & WB & BB1 & BB2 & BB3 & WB & BB1 & BB2 & BB3 & WB & BB1 & BB2 & BB3 & WB & BB1 & BB2 & BB3 & WB & BB1 & BB2 & BB3 \\ 
\midrule
\rowcolor[HTML]{FFFFFF} 
Context-Agnostic & 30.6 & 26.2 & 18.2 & 21.4 & 34.6 & 29.4 & 24.2 & 24.4 & 34.8 & 28.8 & 20.4 & 23.4 & 37.6 & 24.8 & 16.8 & 19.4 & 29.2 & 15.8 & 10.2 & 13.0 \\
\rowcolor[HTML]{FFFFFF} 
\cellcolor[HTML]{E0E0E0}ZQA & 90.4 & 47.8 & 33.0 & 43.4 & 91.8 & 48.4 & 32.2 & 40.0 & 88.6 & 48.6 & 33.0 & 41.8 & \textbf{86.6} & 39.4 & 25.6 & 35.0 & 64.4 & 23.8 & 13.4 & 21.6 \\
\rowcolor[HTML]{FFFFFF} 
\cellcolor[HTML]{C0C0C0}ZQA-PSPM & \textbf{92.2} & \textbf{51.0} & \textbf{34.2} & \textbf{45.0} & \textbf{92.4} & \textbf{52.8} & \textbf{34.4} & \textbf{44.0} & \textbf{92.8} & \textbf{48.8} & \textbf{35.2} & \textbf{42.0} & 86.2 & \textbf{42.8} & \textbf{27.2} & \textbf{37.6} & \textbf{67.2} & \textbf{25.4} & \textbf{14.8} & \textbf{23.4} \\ 
\hline \hline
\rowcolor[HTML]{E0E0E0} 
\cellcolor[HTML]{FFFFFF}Few-Query 0 & 63.0 & 35.6 & 25.8 & 34.0 & 64.2 & 37.0 & 27.0 & 32.2 & 63.2 & 37.0 & 25.2 & 33.0 & 62.2 & 33.2 & 23.2 & 32.2 & 44.0 & 19.0 & 11.0 & 19.6 \\
\rowcolor[HTML]{E0E0E0} 
\cellcolor[HTML]{FFFFFF}Few-Query 1 & 66.8 & 43.6 & 32.4 & 41.0 & 68.2 & 43.6 & \cellcolor[HTML]{C0C0C0}32.4 & 39.2 & 67.8 & 43.8 & 32.0 & 39.4 & 68.0 & \cellcolor[HTML]{C0C0C0}42.8 & \cellcolor[HTML]{FFFFFF}28.8 & \cellcolor[HTML]{FFFFFF}38.2 & 58.6 & \cellcolor[HTML]{FFFFFF}28.0 & \cellcolor[HTML]{FFFFFF}17.8 & \cellcolor[HTML]{FFFFFF}27.4 \\
\rowcolor[HTML]{FFFFFF} 
Few-Query 2 & \cellcolor[HTML]{E0E0E0}77.4 & 51.8 & 38.2 & 47.8 & \cellcolor[HTML]{E0E0E0}76.8 & \cellcolor[HTML]{C0C0C0}51.4 & 37.6 & 47.2 & \cellcolor[HTML]{E0E0E0}78.2 & 51.4 & 37.4 & 45.6 & \cellcolor[HTML]{E0E0E0}76.6 & 47.4 & 33.2 & 43.0 & 67.6 & 33.8 & 21.4 & 33.0 \\
\rowcolor[HTML]{FFFFFF} 
Few-Query 3 & \cellcolor[HTML]{E0E0E0}87.8 & 57.8 & 45.0 & 55.2 & \cellcolor[HTML]{E0E0E0}87.4 & 59.4 & 43.0 & 54.0 & \cellcolor[HTML]{E0E0E0}85.4 & 56.4 & 42.8 & 50.4 & \cellcolor[HTML]{E0E0E0}85.8 & 53.0 & 36.8 & 50.2 & 75.4 & 36.4 & 24.0 & 34.6 \\
\rowcolor[HTML]{FFFFFF} 
Few-Query 4 & 93.4 & 61.0 & 47.4 & 59.4 & 93.4 & 63.6 & 47.0 & 58.8 & \cellcolor[HTML]{C0C0C0}91.4 & 60.6 & 45.4 & 55.8 & 92.4 & 57.4 & 39.8 & 54.0 & 80.8 & 40.2 & 25.8 & 37.6 \\
\rowcolor[HTML]{FFFFFF} 
Few-Query 5 & 96.8 & 64.8 & 49.6 & 61.8 & 97.0 & 67.0 & 49.6 & 61.0 & 95.8 & 63.2 & 47.0 & 58.0 & 95.2 & 59.6 & 41.8 & 57.0 & 84.6 & 41.6 & 27.2 & 39.6 \\
\bottomrule
\end{tabular}}
\end{table*}

%% file: discussion.tex
\section{Discussion}

\PP{Limitations and Future Work:} \label{sec:limitation}
We have assumed that the data distribution at the victim system is known.
We also assume that the context is consistent across surrogate and victim systems. In practice, this is rarely the case. The attacker and the victim system may have different data distributions, overlapping but non-identical label sets, and thus similar but non-identical context models. A useful avenue for future work is to introduce controlled discrepancies between the distributions, context models, and label sets and examine their effect on the fooling rate of the ZQA attack.

While co-occurrence is a fundamental notion of context, it does not capture key properties such as relative size and location of the objects or the relationship between the object and the background. Extending the ZQA attack to more sophisticated context models is a topic for future research. Furthermore, evaluating the ZQA in situations where the attacker uses a different notion of context than the victim system -- e.g., attacker uses semantic context, detector uses context learned from pixels -- would help researchers understand the broader applicability of this work.

Another limitation is that it is expensive to precompute the PSPM. Unlike the context graph, which only depends on the dataset, PSPM is a function of the dataset, attack model and perturbation level. We need to measure the attack success rate for each surrogate model at each perturbation level, and for all possible perturbations of a given object.

\PP{Potential negative societal impact and mitigation:}
This paper, as any other work that investigates an attack method, may be used maliciously to generate attacks against victim systems. Our goal with this work is to reduce technical surprise, and to fuel the development of defenses against powerful attacks. This work already makes the case for using a context-aware detector to thwart simple attacks. To  thwart ZQA attacks described in this paper, the detector can attempt to constantly update its training data and label sets, and to develop increasingly sophisticated context models.

%% file: supp.tex
\newpage
\clearpage
\label{sec:appendix}
\noindent \begin{center} {\large  \textbf{Supplementary Material}} \end{center}

\makeatletter
\def\mysequence#1{\expandafter\@mysequence\csname c@#1\endcsname}
\def\@mysequence#1{%
  \ifcase#1\or -\or - \or -\or -\or -\or -\or A\or B\or C\or D\else\@ctrerr\fi}
\makeatother
\renewcommand\thesection{\mysequence{section}}

In this supplementary material, we provide the experimental results on COCO dataset in \Cref{sec:coco-table}. In \Cref{sec:jpeg-defense}, we show that JPEG defense, one of the most common context-agnostic defense methods, fails against our proposed attack. We also include actual images showing region proposals with detected objects for the zero-query attack, the context-agnostic attack, and the few-query attack. The aim is to demonstrate the visual appearance of the attacked scenes, so that they can ascertain the subjective visual quality of the perturbed scenes, and see examples of cases in which the different attacks succeed or fail in fooling the victim system. \Cref{sec:visualization}.

\section{Experimental results on COCO dataset}
\label{sec:coco-table}
In this section, we repeat the object detection evaluation experiments for the COCO dataset. The models obtained from \texttt{MMDetection} are well trained on COCO2017 training set, and the evaluation results on COCO2017 validation set can be found in \Cref{tab:map-coco}. While the Mean Average Precision (mAP) scores are much lower than those observed for the VOC dataset (Table~\ref{tab:map}), these values are similar to the officially reported numbers in \texttt{MMDetection} repository. This confirms  that the object detection algorithm for the COCO dataset -- a more challenging dataset than VOC -- performs at a level close to the state of the art.
\input{tables/map_coco}
The comparison of ZQA and ZQA-PSPM acting on the COCO dataset against our two baseline schemes is shown in \Cref{tab:compare-frcnn-coco}. As for the VOC dataset, the ZQA attack for the COCO dataset outperforms up to 3 attempts of the Few-Query attack (2 rounds of feedback) in the black-box transfer attack setting. 
\input{tables/compare_frcnn_coco}

\section{Evading context-agnostic defense}
\label{sec:jpeg-defense}
We tested against the commonly used context-agnostic JPEG defense and found that our attack is resilient. Our attack can still outperform up to 5 rounds of few-query attacks under the JPEG compression quality of 95, as shown in \Cref{tab:compare-frcnn-jpeg95}, corresponding to the setting in \Cref{tab:compare-frcnn}.
\input{tables/compare_frcnn_jpeg95}

\section{Visualization of sample images}
\label{sec:visualization}

In this section, we provide visual examples of scenes before and after perturbation. In doing so, we compare the zero-query scheme, the context-agnostic attack, and the few-query scheme that we developed to benchmark performance. All the results are for a transfer setting, i.e., the attacker creates the perturbations on a surrogate model which is different from the classification model used by the victim system. All the images are generated for the case in which the attacker's perturbation is made using a Faster R-CNN network, while the victim system system uses a RetinaNet model. The perturbation budget used to implement the evasion attack is $\epsilon = 10$.

\begin{figure*}[!ht]
    \centering
    \includegraphics[width=0.85\linewidth]{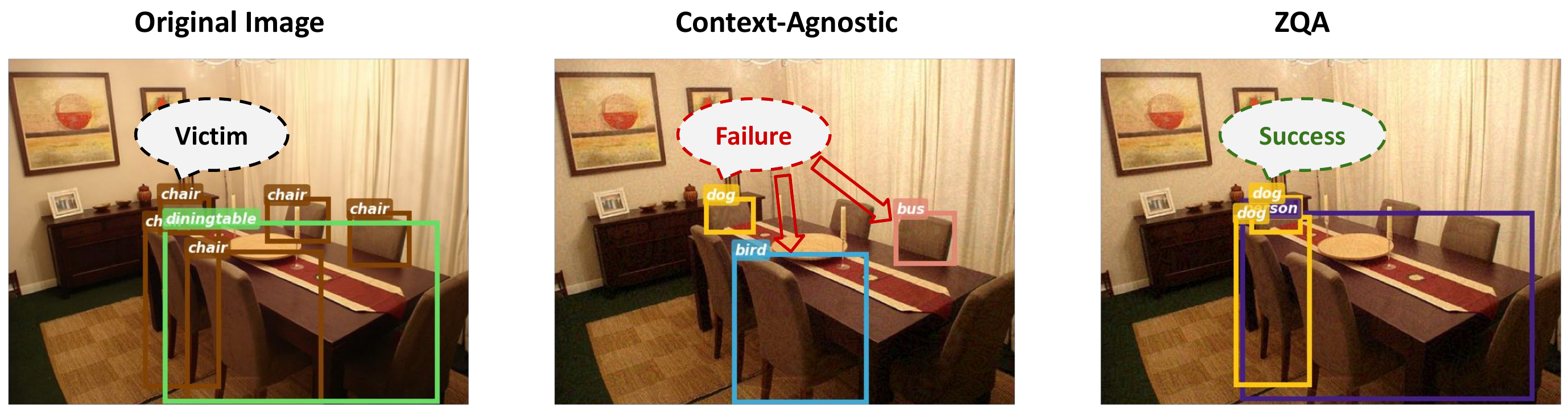}
    \caption{Detections on one original image and images perturbed by the context-agnostic attack and ZQA attack. The goal is to perturb the victim object, which is a chair on the top-left, to a dog. In the transfer attack, both the context-agnostic attack and ZQA attack successfully perturbs the chair to dog, along with some perturbations of surrounding objects. Even though context-agnostic attack is successful in perturbing victim to target, the attack still fails because the surrounding objects (bus and bird) are not context consistent according to the co-occurrence graph.}
    \label{fig:supp-fig1}
\end{figure*}

\Cref{fig:supp-fig1} provides an example in which the context-agnostic attack successfully perturbs the individual objects:  chair $\rightarrow$ dog, chair $\rightarrow$ bus and chair $\rightarrow$ bird. However, the resulting list of detected objects (dog, bus, bird) is context-inconsistent according to the co-occurrence matrix. Thus, the attack is detected. In contrast, the ZQA attack perturbs the objects as follows: chair $\rightarrow$ dog, second chair $\rightarrow$ second dog, dining table $\rightarrow$ person. The list of detected objects (dog, dog, person) is context-consistent, which fools the detector. This shows the basic use case of our context-aware approach.

\begin{figure*}
    \centering
    \includegraphics[width=\linewidth]{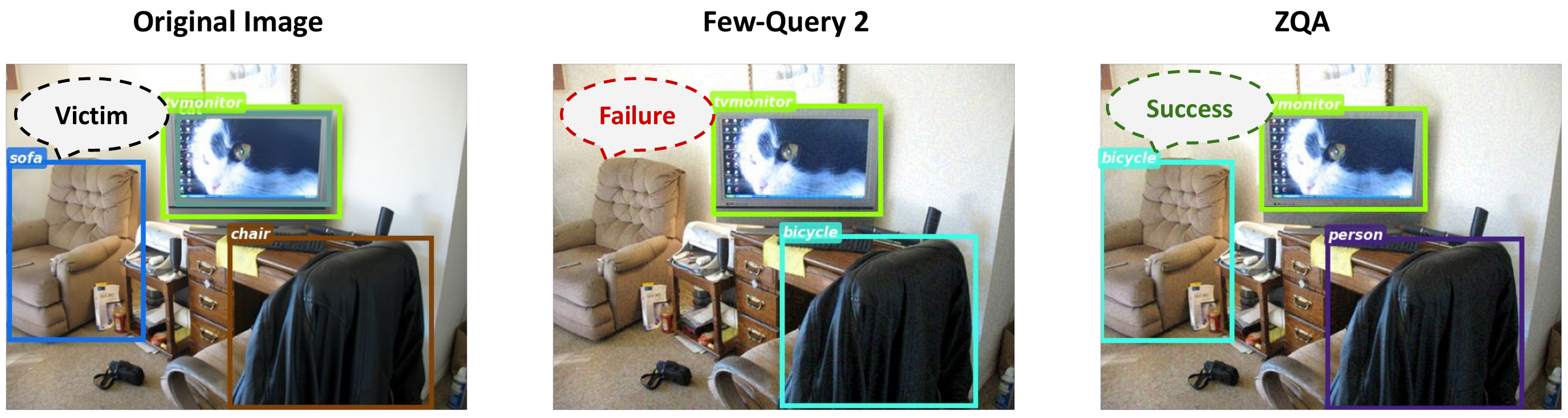}
    \caption{Detections on one original image and images perturbed by the few-query attack and the ZQA attack. The goal is to perturb the victim sofa to a target bicycle. Few-Query attack, building on 2 previous queries, perturbs the sofa to bicycle and the chair to bicycle as well. The TV monitor is not perturbed as it is context consistent. However, the attack failed to transfer to the victim model, in face, not detecting the sofa as a foreground object. Thus, the few-query attack fails. The ZQA attack additionally perturbs the chair to person. Since bicycle, person and TV monitor are all detected and are context-consistent, the attack successfully transfers.}
    \label{fig:supp-fig2}
\end{figure*}

\Cref{fig:supp-fig2} provides an example in which the few-query attack has perturbed the main victim object (sofa $\rightarrow$ bicycle), as well as one other helper object (chair $\rightarrow$ bicycle) in the scene. However, the attack fails because the victim system's detector does not detect the main victim object and relegates it to the background. In contrast, the ZQA attack, with the help of the perturbation success probability matrix (PSPM), chooses object perturbations that are most likely to succeed in a single attempt, i.e., sofa $\rightarrow$ bicycle, chair $\rightarrow$ person, and leaves the TV monitor unchanged. The perturbation applied to the sofa object is sufficient for it to be detected and misclassified as a bicycle. This attack is context-consistent by construction, and successfully fools the detector. We remark here that the vanishing effect scene above is not unique to the few-query attack. Indeed, evasion attacks which involve perturbing the entire scene while attempting to attack individual objects in the scene are susceptible to the vanishing effect. This occurs when the scene perturbation, constrained by the budget $\epsilon$, is such that it causes one or more objects in the scene to not be detected. As expected, we observe this effect more often at lower perturbation budgets.

\begin{figure*}
    \centering
    \includegraphics[width=\linewidth]{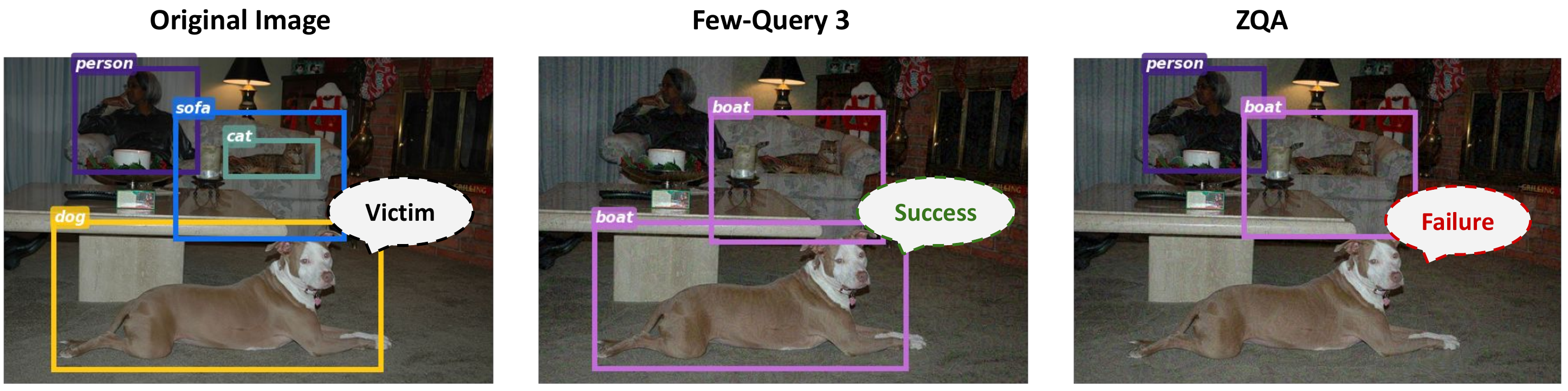}
    \caption{Detections on one original image and images perturbed by few-query attack and ZQA attack. The goal is to perturb the dog to a boat. The few-query attack, building on 3 previous queries, perturbs two objects to boats, and causes the person and the cat to vanish. The result is context-consistent and meets the desired goal. On the other hand, the ZQA attack leaves the person unchanged, perturbs the sofa to a boat, but causes the intended victim object (dog) and another object (cat) to vanish. Even though person and boat are context-consistent in the perturbed scene, the ZQA attack has failed because the intended victim object has vanished.}
    \label{fig:supp-fig3}
\end{figure*}

\Cref{fig:supp-fig3} shows that, given more rounds of feedback, the few-query detector eventually gets enough information about the detector's decisions, and is able to perturb a large number of objects, thereby fooling the detector. The attack attempts to make the following changes: dog $\rightarrow$ boat, sofa $\rightarrow$ boat, cat $\rightarrow$ boat, person $\rightarrow$ boat. The victim system misclassifies the dog and the sofa as boats, but does not detect the person and the cat. Even with the vanishing artifact, we deem the few-query attack successful because it has successfully perturbed the victim object (dog $\rightarrow$ boat) and it has ensured that the detected objects form a context-consistent list. On the other hand, the ZQA attack intends to leave the person unchanged, while changing the other objects to boats. This attack fails because, at the given perturbation level $\epsilon = 10$, the attack left the person unchanged, altered the sofa to the boat, but caused the cat and the dog vanish into the background. This is a failed attack because the main objective of misclassifying the victim object, i.e., dog $\rightarrow$ boat, was not fulfilled. This shows that the few-query approach -- given multiple attempts to enhance the attack -- will eventually overwhelm the proposed ZQA attack which is only allowed a single attempt. One disadvantage of the few query-attack, as noted earlier, is that it requires access to the victim system's communication, thus exposing the attacker to the risk of being discovered. The ZQA attack does not have this limitation.

\begin{figure*}
    \centering
    \includegraphics[width=0.75\linewidth]{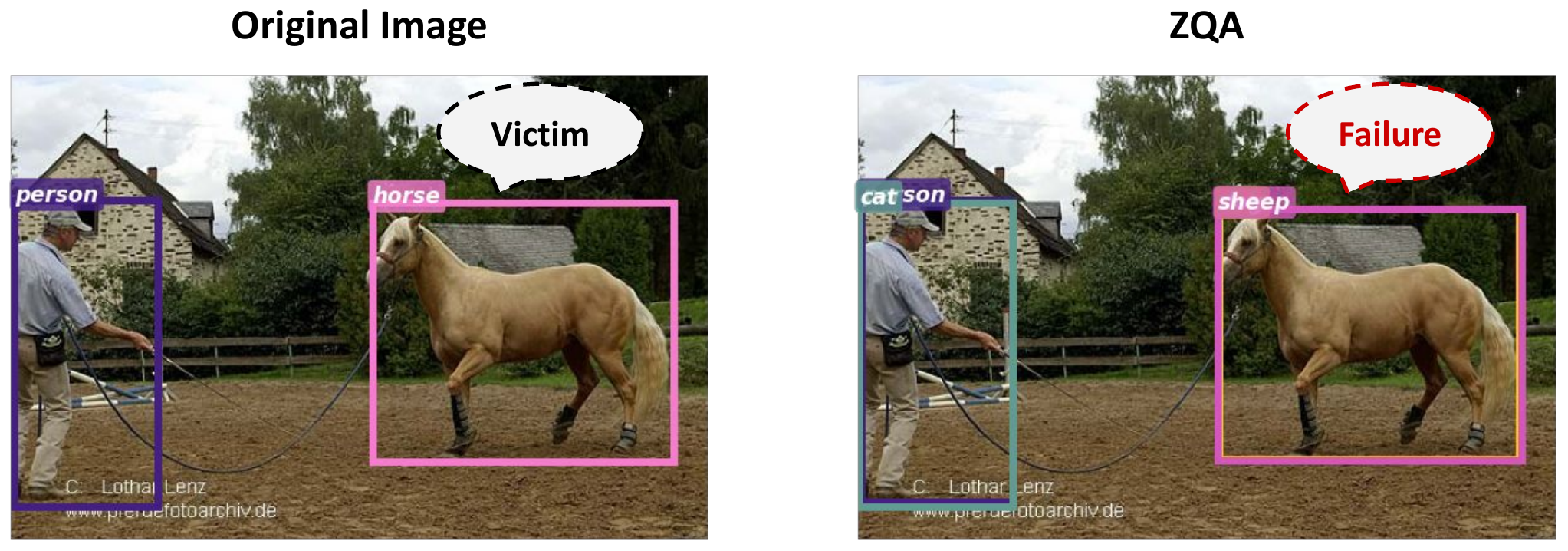}
    \caption{A failure case of ZQA attack. We observe that the perturbation of the victim object (horse $\rightarrow$ cat) does not succeed. Instead, the victim model classifies the perturbed horse as a sheep.}
    \label{fig:supp-fig4}
\end{figure*}

\Cref{fig:supp-fig4} shows one of the failure modes of our approach. (This type of failure is also observed in general perturbation bounded evasion attacks, and in our case, it is also seen in some cases of the few-query attack, and the context-agnostic attack). The goal of the attacker is to perturb the horse to a cat. However, the attack made with the surrogate model does not correctly transfer to the black-box victim model. The detector recognizes the horse as a sheep, which is unintended for our targeted attack.

%% file: tables/map_coco.tex
\begin{table}[h]
\centering
\caption{Mean average precision (mAP) at IOU (intersection over union) threshold 0.5 of different detectors used in our experiments. Models are evaluated on COCO2017 val set. \textbf{Legend:} Faster R-CNN (FRCNN), RetinaNet (Retina), Libra R-CNN (Libra), FoveaBox (Fovea).}
\label{tab:map-coco}
\begin{tabular}{cccccc}
\hline
Model   & FRCNN & Retina & Libra & Fovea \\
mAP@.50 & 38.99\% & 35.13\% & 40.14\% & 45.78\%  \\ \hline
\end{tabular}
\end{table}

%% file: tables/compare_frcnn_coco.tex
\begin{table*}[h]
\centering
\caption{
Follow the setting in \Cref{tab:compare-frcnn} but use 500 images from COCO 2017 test set. 
Fooling rates (\%) of different attack strategies under different $L_{\infty}$ perturbation $\le \epsilon \in \{50,40,30,20,10\}$ are as follows.
}
\label{tab:compare-frcnn-coco}
\scalebox{0.75}{
\begin{tabular}{c|cccccccccccccccccccc}
\toprule
\rowcolor[HTML]{FFFFFF} 
\cellcolor[HTML]{FFFFFF} & \multicolumn{4}{c}{\cellcolor[HTML]{FFFFFF}$\epsilon = 50$} & \multicolumn{4}{c}{\cellcolor[HTML]{FFFFFF}$\epsilon = 40$} & \multicolumn{4}{c}{\cellcolor[HTML]{FFFFFF}$\epsilon = 30$} & \multicolumn{4}{c}{\cellcolor[HTML]{FFFFFF}$\epsilon = 20$} & \multicolumn{4}{c}{\cellcolor[HTML]{FFFFFF}$\epsilon = 10$} \\ \cline{2-21} 
\rowcolor[HTML]{FFFFFF} 
\multirow{-2}{*}{\cellcolor[HTML]{FFFFFF}Method} & WB & BB1 & BB2 & BB3 & WB & BB1 & BB2 & BB3 & WB & BB1 & BB2 & BB3 & WB & BB1 & BB2 & BB3 & WB & BB1 & BB2 & BB3 \\ 
\midrule
\rowcolor[HTML]{FFFFFF} 
Context-Agnostic & 55.2 & 23.8 & 27.2 & 35.2 & 60.0 & 25.8 & 28.0 & 31.2 & 55.4 & 23.4 & 21.6 & 31.8 & 52.2 & 18.8 & 18.6 & 28.6 & 39.6 & 14.2 & 12.4 & 15.8 \\
\rowcolor[HTML]{FFFFFF} 
\cellcolor[HTML]{E0E0E0}ZQA & 82.2 & 29.8 & 35.4 & 43.6 & 82.8 & 30.2 & 35.8 & 43.2 & 81.0 & \textbf{30.4} & 31.0 & 40.0 & 76.0 & 23.8 & 26.4 & \textbf{37.4} & 52.0 & 14.2 & 15.6 & 21.8 \\
\rowcolor[HTML]{FFFFFF} 
\cellcolor[HTML]{C0C0C0}ZQA-PSPM & \textbf{85.0} & \textbf{34.0} & \textbf{38.0} & \textbf{48.0} & \textbf{85.8} & \textbf{32.0} & \textbf{39.6} & \textbf{43.8} & \textbf{82.8} & 29.8 & \textbf{32.6} & \textbf{46.0} & \textbf{79.0} & \textbf{27.2} & \textbf{29.8} & 36.8 & \textbf{58.2} & \textbf{15.6} & \textbf{17.6} & \textbf{25.0} \\ 
\hline \hline
\rowcolor[HTML]{E0E0E0} 
\cellcolor[HTML]{FFFFFF}Few-Query 0 & 71.4 & 27.0 & 24.0 & 37.8 & 73.4 & 25.6 & 24.0 & 35.0 & 68.4 & 21.8 & 18.2 & 34.0 & 63.8 & 21.2 & 17.6 & 28.4 & 47.2 & 13.2 & 9.6 & 18.4 \\
\rowcolor[HTML]{E0E0E0} 
\cellcolor[HTML]{FFFFFF}Few-Query 1 & 80.0 & \cellcolor[HTML]{FFFFFF}34.2 & 34.2 & \cellcolor[HTML]{C0C0C0}46.6 & 80.0 & \cellcolor[HTML]{FFFFFF}33.6 & 34.2 & \cellcolor[HTML]{FFFFFF}44.0 & 79.0 & 29.8 & 27.2 & \cellcolor[HTML]{C0C0C0}44.4 & 72.6 & \cellcolor[HTML]{C0C0C0}26.0 & 24.6 & 36.8 & \cellcolor[HTML]{C0C0C0}56.4 & \cellcolor[HTML]{FFFFFF}19.2 & 15.4 & \cellcolor[HTML]{FFFFFF}26.0 \\
\rowcolor[HTML]{FFFFFF} 
Few-Query 2 & \cellcolor[HTML]{C0C0C0}83.4 & 37.8 & 41.4 & 51.4 & \cellcolor[HTML]{C0C0C0}84.0 & 39.4 & \cellcolor[HTML]{C0C0C0}39.0 & 50.4 & 84.2 & 34.2 & 33.2 & 49.8 & 79.2 & 31.4 & 30.2 & 43.6 & 62.4 & 21.8 & 19.0 & 30.6 \\
\rowcolor[HTML]{FFFFFF} 
Few-Query 3 & 86.2 & 40.6 & 46.2 & 55.2 & 86.8 & 41.8 & 42.2 & 54.6 & 86.2 & 36.6 & 39.2 & 53.6 & 81.6 & 33.2 & 34.8 & 46.6 & 66.8 & 23.6 & 21.2 & 34.0 \\
\rowcolor[HTML]{FFFFFF} 
Few-Query 4 & 88.0 & 42.8 & 48.0 & 57.8 & 89.8 & 42.8 & 45.4 & 56.6 & 87.8 & 37.8 & 42.0 & 55.4 & 84.4 & 35.2 & 38.0 & 49.2 & 69.0 & 24.2 & 23.0 & 36.2 \\
\rowcolor[HTML]{FFFFFF} 
Few-Query 5 & 89.2 & 45.0 & 52.4 & 59.2 & 92.0 & 44.4 & 48.0 & 57.8 & 89.8 & 39.6 & 45.0 & 57.6 & 85.8 & 36.0 & 40.6 & 51.6 & 71.4 & 25.6 & 25.4 & 38.2 \\
\bottomrule
\end{tabular}}
\end{table*}

%% file: tables/compare_frcnn_jpeg95.tex
\begin{table*}[ht]
\centering
\caption{
Follow the setting in \Cref{tab:compare-frcnn} but under the JPEG compression quality of 95.
Fooling rates (\%) of different attack strategies under different $L_{\infty}$ perturbation $\le \epsilon \in \{50,40,30,20,10\}$ are as follows.}
\label{tab:compare-frcnn-jpeg95}
\scalebox{0.75}{
\begin{tabular}{c|cccccccccccccccccccc}
\toprule
\rowcolor[HTML]{FFFFFF} 
\cellcolor[HTML]{FFFFFF} & \multicolumn{4}{c}{\cellcolor[HTML]{FFFFFF}$\epsilon = 50$} & \multicolumn{4}{c}{\cellcolor[HTML]{FFFFFF}$\epsilon = 40$} & \multicolumn{4}{c}{\cellcolor[HTML]{FFFFFF}$\epsilon = 30$} & \multicolumn{4}{c}{\cellcolor[HTML]{FFFFFF}$\epsilon = 20$} & \multicolumn{4}{c}{\cellcolor[HTML]{FFFFFF}$\epsilon = 10$} \\ \cline{2-21} 
\rowcolor[HTML]{FFFFFF} 
\multirow{-2}{*}{\cellcolor[HTML]{FFFFFF}Method} & WB & BB1 & BB2 & BB3 & WB & BB1 & BB2 & BB3 & WB & BB1 & BB2 & BB3 & WB & BB1 & BB2 & BB3 & WB & BB1 & BB2 & BB3 \\ 
\midrule
Context-Agnostic & 34.6 & 26.4 & 30.0 & 25.6 & 33.4 & 23.6 & 25.0 & 26.0 & 34.4 & 26.4 & 28.8 & 27.0 & 38.4 & 24.0 & 23.6 & 25.8 & 28.2 & 9.4 & 11.0 & 14.8 \\
\cellcolor[HTML]{E0E0E0}ZQA & 88.2 & 41.4 & 49.4 & 51.4 & 86.8 & 40.0 & 47.8 & 47.0 & 88.2 & 41.4 & 49.6 & 47.4 & 82.4 & 35.6 & 40.6 & 42.2 & 49.6 & 14.2 & \textbf{16.8} & 20.0 \\
\cellcolor[HTML]{C0C0C0}ZQA-PSPM & \textbf{89.2} & \textbf{42.8} & \textbf{50.2} & \textbf{53.8} & \textbf{90.2} & \textbf{41.2} & \textbf{48.6} & \textbf{49.8} & \textbf{92.8} & \textbf{44.2} & \textbf{52.2} & \textbf{51.2} & \textbf{83.6} & \textbf{36.4} & \textbf{42.0} & \textbf{44.2} & \textbf{55.8} & \textbf{15.6} & 15.2 & \textbf{21.4} \\ 
\hline \hline
Few-Query 0 & \cellcolor[HTML]{E0E0E0}62.2 & \cellcolor[HTML]{E0E0E0}28.2 & \cellcolor[HTML]{E0E0E0}28.6 & \cellcolor[HTML]{E0E0E0}36.0 & \cellcolor[HTML]{E0E0E0}62.8 & \cellcolor[HTML]{E0E0E0}26.8 & \cellcolor[HTML]{E0E0E0}28.6 & \cellcolor[HTML]{E0E0E0}33.6 & \cellcolor[HTML]{E0E0E0}64.4 & \cellcolor[HTML]{E0E0E0}28.8 & \cellcolor[HTML]{E0E0E0}30.6 & \cellcolor[HTML]{E0E0E0}33.0 & \cellcolor[HTML]{E0E0E0}60.6 & \cellcolor[HTML]{E0E0E0}23.6 & \cellcolor[HTML]{E0E0E0}24.8 & \cellcolor[HTML]{E0E0E0}31.2 & \cellcolor[HTML]{E0E0E0}39.0 & \cellcolor[HTML]{E0E0E0}10.8 & \cellcolor[HTML]{E0E0E0}10.8 & \cellcolor[HTML]{E0E0E0}16.6 \\
Few-Query 1 & \cellcolor[HTML]{E0E0E0}68.0 & \cellcolor[HTML]{E0E0E0}37.2 & \cellcolor[HTML]{E0E0E0}39.6 & \cellcolor[HTML]{E0E0E0}45.6 & \cellcolor[HTML]{E0E0E0}70.4 & \cellcolor[HTML]{E0E0E0}33.2 & \cellcolor[HTML]{E0E0E0}37.8 & \cellcolor[HTML]{E0E0E0}41.8 & \cellcolor[HTML]{E0E0E0}68.8 & \cellcolor[HTML]{E0E0E0}35.6 & \cellcolor[HTML]{E0E0E0}39.6 & \cellcolor[HTML]{E0E0E0}42.8 & \cellcolor[HTML]{E0E0E0}66.8 & \cellcolor[HTML]{E0E0E0}31.4 & \cellcolor[HTML]{E0E0E0}32.4 & \cellcolor[HTML]{E0E0E0}40.0 & \cellcolor[HTML]{E0E0E0}46.2 & 16.6 & \cellcolor[HTML]{E0E0E0}15.2 & 22.6 \\
Few-Query 2 & \cellcolor[HTML]{E0E0E0}78.8 & 44.0 & \cellcolor[HTML]{C0C0C0}50.2 & 55.8 & \cellcolor[HTML]{E0E0E0}78.2 & \cellcolor[HTML]{C0C0C0}40.8 & 49.6 & 52.2 & \cellcolor[HTML]{E0E0E0}76.8 & \cellcolor[HTML]{C0C0C0}42.0 & \cellcolor[HTML]{E0E0E0}48.0 & \cellcolor[HTML]{C0C0C0}50.8 & \cellcolor[HTML]{E0E0E0}76.0 & 40.0 & 42.4 & 47.2 & 56.2 & 20.8 & 19.6 & 28.4 \\
Few-Query 3 & \cellcolor[HTML]{E0E0E0}87.4 & 48.8 & 57.8 & 61.6 & \cellcolor[HTML]{E0E0E0}85.8 & 48.6 & 57.4 & 57.6 & \cellcolor[HTML]{E0E0E0}84.4 & 49.8 & 55.8 & 58.6 & \cellcolor[HTML]{C0C0C0}82.8 & 45.6 & 49.4 & 53.4 & 62.8 & 23.6 & 23.6 & 30.2 \\
Few-Query 4 & 91.0 & 52.6 & 62.4 & 64.4 & \cellcolor[HTML]{C0C0C0}90.2 & 50.8 & 61.6 & 61.8 & \cellcolor[HTML]{C0C0C0}88.8 & 52.4 & 61.0 & 62.8 & 88.2 & 48.8 & 53.0 & 57.6 & 68.2 & 25.8 & 26.8 & 33.0 \\
Few-Query 5 & 93.8 & 55.8 & 66.4 & 66.4 & 94.6 & 53.0 & 65.4 & 65.8 & 94.8 & 55.2 & 64.4 & 67.0 & 90.8 & 50.6 & 55.4 & 60.6 & 71.2 & 28.0 & 28.8 & 34.8 \\
\bottomrule
\end{tabular}}
\end{table*}